\documentclass[runningheads]{llncs}
\usepackage{cite}
\usepackage{graphicx}
\usepackage{hyperref}
\hypersetup{
colorlinks=true,linkcolor=black,
urlcolor=blue
}
\begin{document}
\title{Towards Supervised Extractive Text Summarization via RNN-based Sequence Classification}
\titlerunning{Towards Supervised Extractive Text Summarization via RNN Classification}
%
\author{Eduardo Brito \and
Max L\"ubbering\and
David Biesner \and Lars Patrick Hillebrand \and Christian Bauckhage}
\authorrunning{E. Brito et al.}
%
\institute{Fraunhofer IAIS, Sankt Augustin, Germany \\
Fraunhofer Center for Machine Learning, Germany
\\
B-IT, University of Bonn, Bonn, Germany}
\maketitle              
\begin{abstract}
  This article briefly explains our submitted approach to the DocEng'19 competition on extractive summarization\cite{Lins:2019:DCE:3342558.3351874}. We implemented a recurrent neural network based model that learns to classify whether an article's sentence belongs to the corresponding extractive summary or not. We bypass the lack of large annotated news corpora for extractive summarization by generating extractive summaries from abstractive ones, which are available from the CNN corpus.

\end{abstract}

\section{Introduction}

The DocEng '19 competition focused on automatic extractive text summarization. Participants were provided with a corpus of 50 news articles from the CNN-corpus\cite{Lins:2019:CLT:3342558.3345388}. These articles contained corresponding extractive and abstractive summaries aimed to train and test a system to perform the summarization task. The gold standard summaries contained around 10\% of the original text, with a minimum of 3 sentences. After submission, the methods were tested on a larger test set consisting of 1000 articles randomly chosen from the CNN-corpus.
The limited available training data was one of the major challenges of this competition, which prevented any deep learning approach from being successful if no external corpus was incorporated to the training set.

\section{Approach}

\begin{figure}[t!]
     \centering
        \includegraphics[width=0.66\linewidth]{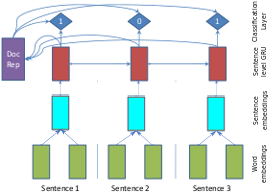}
     \caption{Our RNN-based sequence classifier (based on \cite{nallapati2017summarunner}). All word embeddings from each sentence are averaged to generate a sentence embedding. Sentence embeddings are then used for the bidirectional RNN at sentence level. At the top, the sigmoid activation based classification layer decides whether a sentence is included in the summary based on the content richness of the sentence, its salience with respect to the document and its novelty respect to the accumulated summary representation.}
     \label{fig:architecture}
\end{figure}

Our work is based on the SummaRuNNer model \cite{nallapati2017summarunner}. It consists of a two-layer bi-directional Gated Recurrent Unit (GRU) Recurrent Neural Network (RNN) which treats the summarization problem as a binary sequence classification problem, where each sentence is classified sequentially as sentence to be included or not in the summary. However, we introduced two modifications to the original SummaRuNNer architecture, leading to better results while reducing complexity:
\begin{enumerate}
    \item Our model operates directly on a sentence level (instead of at word level within each sentence). We compute sentence vector representations by means of the the \emph{Flair} library. \cite{akbik2018coling}\footnote{\url{https://github.com/zalandoresearch/flair}}. These sentence embeddings substitute the bottom layer of the SummaRuNNer architecture.
    \item We do not consider the position of each sentence (absolute or relative) for the logistic layer.
\end{enumerate}
The resulting architecture is displayed on Figure \ref{fig:architecture}. Our code to generate extractive summaries according to the instructions established for the competition is publicly available\footnote{\url{https://jira.iais.fraunhofer.de/stash/users/dbiesner/repos/doceng2019_fraunhofer}}.

\section{Data}
In contrast to \cite{nallapati2017summarunner}, we trained our model only on CNN articles from the CNN/Daily Mail corpus\cite{nips15_hermann}. 
Due to the limited number of provided news articles, we automatically annotated a large corpus of CNN articles from which an abstractive summary was available. In a similar approach to \cite{nallapati2017summarunner}, we calculated the ROUGE-1 F1 score between each sentence and its article's abstractive summary. Finally for each article, we sorted the sentences having the highest ROUGE-1 F1 score and picked the top $N=\max(0.1* ||\textrm{sentences}||, 3)$ sentences.

\begin{table}
\centering
  \caption{Evaluation on the 50 labeled news articles provided by the competition organizers}
  \label{tab:evaluation}
  \begin{tabular}{clll}
    \hline
    Score&Precision&Recall& F1\\
    \hline
    Sentence matching gold standard &0.375& 0.357& 0.358\\
    ROUGE-1& 0.384 & 0.206 & 0.261\\
    ROUGE-2 &0.141 & 0.094 & 0.094\\
  \hline
\end{tabular}
\end{table}
\section{Evaluation}
We evaluated our model on the provided labeled CNN news articles with three different metrics: sentences from the generated summary matching the gold standard summary, ROUGE-1 and ROUGE-2. The achieved scores with our trained model after 20 epochs are displayed on Table \ref{tab:evaluation}.
\section{Conclusion}
Our approach achieved the second best performance among the compared methods in the competition, although the F1-score difference between both approaches is not statistically significant\cite{Lins:2019:DCE:3342558.3351874}. Additionally, the performance of these approaches is hardly better than some of the "traditional algorithms" that were presented as baselines, which are  much simpler than ours. Moreover, the real value of the different approaches on the various use cases of automatic text summarization cannot be covered with the current evaluation since the valuable properties of the summaries vary depending on the use case. For instance, coherence is important if the summary will be read by a final user while it is not if the summary is "just" a preprocessing step within an indexing pipeline. Therefore, it would be interesting to assess the different techniques on several downstream tasks to obtain a better overview about which algorithms are most suitable.
%
%
%
 \bibliographystyle{splncs04}
 \bibliography{mybibliography}

\end{document}